\documentclass[journal]{IEEEtran}
\usepackage{times}

\usepackage[numbers]{natbib}
\usepackage{multicol}
\usepackage[bookmarks=true]{hyperref}
\usepackage[pdftex]{graphicx}
\usepackage{amsfonts}
\usepackage{amsmath}
\usepackage{float}
\usepackage{subfigure}
\usepackage{comment}
\usepackage{color}

\begin{document}
                  

\title{Efficient Model Learning for Human-Robot Collaborative Tasks}



\author{Stefanos Nikolaidis, Keren Gu, Ramya Ramakrishnan, and Julie Shah
\thanks{Stefanos Nikolaidis (corresponding author),  Keren Gu, Ramya Ramakrishnan, and Julie Shah are with the  Massachusetts Institute of Technology, Cambridge, MA, 02139 USA. Email: snikol@mit.edu, kgu@mit.edu, ramyaram@mit.edu, julie\_a\_shah@csail.mit.edu}}



%

\maketitle

\begin{abstract}

We present a framework for learning human user models from joint-action demonstrations that enables the robot to compute a robust policy for a collaborative task with a human.
The learning takes place completely automatically, without any human intervention.  First, we describe the clustering of demonstrated action sequences into different human types using an unsupervised learning algorithm. These demonstrated sequences are also used by the robot to learn a reward function that is representative for each type, through the employment of an inverse reinforcement learning algorithm. The learned model is then used as part of a Mixed Observability Markov Decision Process formulation, wherein the human type is a partially observable variable. With this framework, we can infer, either offline or online, the human type of a new user that was not included in the training set, and can compute a policy for the robot that will be aligned to the preference of this new user and will be robust to deviations of the human actions from prior demonstrations. Finally we validate the approach using data collected in human subject experiments, and conduct proof-of-concept demonstrations in which a person performs a collaborative task with a small industrial robot. 


\end{abstract}

\IEEEpeerreviewmaketitle

\section{Introduction}
The development of new industrial robotic systems that operate in the same physical space as people highlights the emerging need for robots that can integrate seamlessly into human group dynamics by adapting to the personalized style of human teammates. This adaptation requires learning a statistical model of human behavior and integrating this model into the decision-making algorithm of the robot in a principled way. 

We present a framework for learning human user models from joint-action demonstrations that enables the robot to compute a robust policy for a collaborative task with a human, assuming access to demonstrations of human teams working on the task.
The learning takes place completely automatically, without any human intervention. Additionally, the robustness of the action selection mechanism of the robot is compared to previous model-learning algorithms in the ability to function despite increasing deviations of human actions from previously demonstrated behavior.  

This work is based on our observation that even with a large number of different human preferences, the actual high-level strategies followed by humans working in teams are generally limited in number. We use this insight to denote the preference of a human team member for the actions of his robotic teammate as a partially observable variable in a Mixed Observability Markov Decision Process \cite{ong2010planning}, and constrain its value to a limited set of possible assignments. We chose the MOMDP formulation because the number of observable variables for human-robot collaborative tasks in a manufacturing setting is much larger than that of partially observable variables. We define as ``human type" the preference the human has for a subset of the task-related actions taken by the robot during a collaborative task. Denoting the human preference for the actions of his partner as a hidden variable naturally models human collaboration, since the intentions of the participants can never be directly observed during training, and must be inferred through interaction and observation. 


First, we describe the clustering of demonstrated action sequences into different human types using an unsupervised learning algorithm. These demonstrated sequences are then used by the robot to learn a reward function that is representative for each type, through the employment of an inverse reinforcement learning algorithm. The learned model is then used as part of a MOMDP formulation, wherein the human type is a partially observable variable. With this framework, we can infer, either offline or online, the human type of a new user that was not included in the training set, and can compute a policy for the robot that will be aligned to the preference of this new user and will be robust to deviations of the human actions from prior demonstrations. Finally we validate the approach using data collected in human subject experiments, and conduct proof-of-concept demonstrations in which a person performs a collaborative task with a small industrial robot. 

\section{Relevant Work}

For a robot to learn a human model, a human expert is typically required to explicitly teach the robot a skill or specific task  \cite{Argall2009,AtkesonS97,Abbeel04,Nicolescu03naturalmethods,Chernova2008,AkgunCYT12}. In a manufacturing setting, a large part of work is manually performed by humans. When performing manual work, people develop their own personalized style of completing a task, although some aspects of the task may be well-defined. In this work, demonstrations of human teams executing a task are used to automatically learn human types in an unsupervised fashion. The data from each cluster is then inputted to an inverse reinforcement learning algorithm. In the context of control theory, this problem is known as Inverse Optimal Control, originally posed by Kalman and solved in \cite{BEFB94}. There have been a number of inverse reinforcement learning methods developed, many of which use a weighted-features representation for the unknown reward function. We follow the approach of Abbeel and Ng \cite{Abbeel04}, and solve a quadratic program iteratively to find feature weights that attempt to match the expected feature counts of the resulting policy with those of the expert demonstrations. Other approaches involve finding a weight vector that explains the expert demonstrations by optimizing the margin between competing explanations. There have also been game-theoretic approaches \cite{syed2007game, Ziebart2011} that aim to model multi-agent behavior. Recent state-of-the-art imitation learning algorithms \cite{Kim-RSS-13} have been shown to preserve theoretical guarantees of performance, while improving the safety of the exploration.
Through human demonstrations, our framework learns a number of different human types and a reward function for each type, and uses these as part of a MOMDP formulation. 

Related approaches to learning user models include natural language interaction with a robot wheelchair \cite{Doshi07HRI}, where a user model is learned simultaneously with a dialog manager policy. The human interacts with the system by issuing verbal commands, and offering a scalar reward after each robot action. The model is encoded in the transition functions, observation functions and rewards of a Partially Observable Markov Decision Process framework. The system assumes that the model parameters are initially uncertain, and improves the model through interaction. Rather than learning a new model for each human user, which can be tedious and time-consuming, we use demonstrations by human teams to infer some ``dominant" human types and then associate each new user to a new type. 

In prior work involving pursuit games, researchers employed an empirical approach in which an agent plans using Monte-Carlo Tree Search with a set of known models of possible teammates, which are then used to generate action likelihoods to infer teammate types from observed behavior \cite{AAMAS11barrett}. Rather than using a fixed set of known models, we estimate the models automatically from training data. Another approach \cite{wang2012probabilistic} learned non-parametric Bayesian models of human movements from demonstration to train robots to play table-tennis with human partners. These models required a set of labeled demonstrations, whereas we cluster unlabeled actions sequences into discrete human types using unsupervised learning algorithms. 

Recent work has also inferred human intentions in collaborative tasks for game AI applications.  \cite{nguyen2011capir} focused on inferring the intentions of a human player, allowing a Non-Player Character (NPC) to assist the human. They developed the CAPIR framework, in which a task is deconstructed into subtasks, each of which is computationally tractable and modeled by a Markov Decision Process. Alternatively,~\cite{macindoe2012pomcop} proposed the Partially Observable Monte-Carlo cooperative planning system, in which human intention is inferred for a cops-and-robbers turn-based game. The algorithm uses a black-box simulator to generate human actions, and interfaces it with a Monte-Carlo planner \cite{silver2010monte}. In both works, the model of the human type is assumed to be known beforehand.

Partially Observable Markov Decision Process models have been used to infer human intention during driving tasks~\cite{broz2011designing} as well. There the hidden variable is the intention of a human for their own actions. The user model is represented by the transition matrix of a POMDP and is learned through task-specific action-rules.
In our framework, none of the learning steps require task-specific rules. Alternative POMDP models of multi-agent collaboration have been used for interactive assistant applications~\cite{fern2010computational}. Recently, the mixed observability predictive state representation framework (MO-PSR) \cite{ong2013mixed} has been shown to learn accurate models of mixed observability systems directly from observable quantities, but has not been verified yet for task planning applications. The MOMDP formulation \cite{ong2010planning} has been shown to achieve significant computational efficiency, and has been used in motion planning applications \cite{bandyopadhyay2013intention}, with uncertainty about the intention of the human over their own actions. In the aforementioned work, the reward structure of the task is assumed to be known. 
In our work, the reward function that corresponds to each human type is learned automatically from unlabeled demonstrations. 

In summary, the proposed framework makes the following contributions:
\begin{itemize}
\item It enables the rapid estimation of a human user model, which can be done either offline or online, through the a priori unsupervised learning of a set of "dominant" models. This differs from previous approaches \cite{Doshi07HRI} that start with uncertain model parameters and learn them through interaction. Such approaches do not have the limitation of a fixed set of available models, however learning a good model requires a very large amount of data, which can be an issue when using them for practical applications.
\item It uses a MOMDP formulation to compute personalized policies for the robot that take uncertainty about the human type into consideration. Similar MOMDP formulations have been used in prior work  \cite{ong2010planning}, \cite{bandyopadhyay2013intention}, but with the reward structure assumed to be known. Research on POMDP formulations for collaborative tasks in game AI applications \cite{nguyen2011capir, macindoe2012pomcop, silver2010monte} also assumed a known human model. We present a pipeline to automatically learn the reward function of the MOMDP through unsupervised learning and inverse reinforcement learning. 
\item  It presents a MOMDP formulation with a human type as the partially observable variable, and the reward function as a function of the human type. This allows the computation of a policy that is in accordance with the preference of the human teammate over what the robot should do. Previous partially observable formalisms \cite{ong2010planning, bandyopadhyay2013intention, broz2011designing, fern2010computational, nguyen2011capir, macindoe2012pomcop} in assistive or collaborative tasks represented the preference or intention of the human for their own actions, rather than those of the robot, as the partially observable variable. 
\item It is validated using data from actual human subject experiments to show that the learned MOMDP policies perform similarly to policies obtained from a domain expert using a hand-coded model, and significantly more robustly than previous algorithms for human-robot collaborative tasks that reason in state-histories \cite{Nikolaidis2013}.  

\end{itemize}
We describe the proposed framework in the next section.

\section{Method}
\begin{figure}[h]
\centering
 {\includegraphics[width=1.0\linewidth]{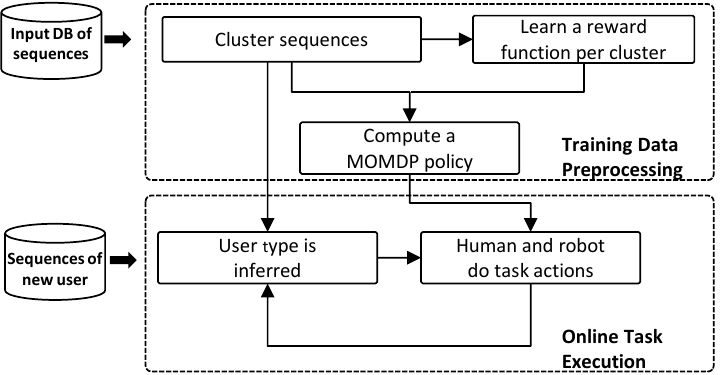}}\hfil
 \caption{Framework flowchart }

 \label{fig:flowchart}
\end{figure}

Our proposed framework has two main stages, as shown in Figure~\ref{fig:flowchart}. The training data is preprocessed in the first stage. In the second stage, the robot infers the personalized style of a new human teammate and executes its role in the task according to the preference of this teammate. 


When a robot is introduced to work with a new human worker, it needs to infer the human type and choose actions aligned to the preference of that human. Additionally, the robot should reason over the uncertainty on the type of the human. The first stage of our framework assumes access to a set of demonstrated sequences of actions from human teams working together on a collaborative task, and uses an unsupervised learning algorithm to cluster the data into dominating human types. The cluster indices serve as the values of a partially observable variable denoting human type, in a Mixed-Observability Markov Decision Process. Our framework then learns a reward function for each human type, which represents the preference of a human of the given type on a subset of task-related robot actions. Finally, the framework computes an approximately optimal policy for the robot that reasons over the uncertainty on the human type and maximizes the expected accumulated reward. 

In the second stage, a new human subject is asked to execute the collaborative task with the robot. The human is first instructed to demonstrate a few sequences of human and robot actions. A belief about his type is then computed according to the likelihood of the human sequences belonging to each cluster. Alternatively, if the human actions are informative of his type \textemdash his preference for the actions of the robot \textemdash the human type can be estimated online. The robot then executes the action based on the computed policy of the MOMDP, based on the current belief of the human type, at each time step.

In the following section, we describe the first block of the proposed framework: finding the number of dominating human types in a collaborative task by clustering the demonstrated sequences.

\section{Clustering of Human Types} \label{sec:clustering}

To improve a robot's ability to adapt to human preferences, we first try to find human preferences using an unsupervised clustering approach. In this problem, we have a data set $D=x_{1},...,x_{n}$, where each $x_{i}$ is a demonstrated sequence of alternating discrete human and robot actions. The robot actions are those that the human would like the robot to take. We can determine these actions, for example, by observing how two humans work together. The goal is to find the number of human types, $k$,  within this data and the assignment of each sequence of actions $x_{i}$ to a type. 

Previous work has approached this problem of clustering sequential data using various methods. Murphy and Martin \cite{murphy2003mixtures} clustered ranking or ordinal data through Expectation Maximization (EM) by learning distance-based models that had two parameters: a central ranking and a precision parameter. The distance between rankings was defined using Kendall's, Spearman's and Cayley's distances, as specified in \cite{marden1995distances}. To select the best model, Bayesian information criterion (BIC) and integrated complete likelihood (ICL) were used. In another work, J{\"a}{\"a}skinen \cite{jaaskinen2013bayesian} clustered DNA sequences modeled as Markov chains using a Dirichlet process prior over the partitions. A greedy search of joining and splitting partitions was used to determine the number of clusters, and EM was used to learn transition probability matrices and to correctly assign sequences to clusters.

In solving our clustering problem, we chose to use a hybrid approach of these two methods. Similar to \cite{jaaskinen2013bayesian}, our framework learns transition matrices between human and robot actions using EM because this provides information about how the human will act based on the actions of the robot, and vice versa. However, we use a uniform prior distribution over the partitions, rather than the Dirichlet process prior \cite{jaaskinen2013bayesian}, as this was sufficient for our task. We use BIC to find the ideal value of $k$, as done in \cite{murphy2003mixtures}, rather than the greedy approach in \cite{jaaskinen2013bayesian}, due to the small number of possible values of $k$. 
Again, we based this on the observation that, even in complex tasks, the actual high-level strategies followed by humans working in teams are usually few in number. 

We begin by using a hard variant of EM, similar to \cite{jaaskinen2013bayesian}, to cluster the data into a set of human preferences. In the algorithm, we represent each preference or cluster by a transition matrix of size $|A|$ x $|A|$, where $|A|$ is the size of the action space, $A = \{A_{r}, A_{h}\}$, which includes both robot actions $A_{r}$ and human actions $A_{h}$. Since the data consists of a sequence of actions in which the human and robot take turns, the transition matrix encodes information about how the human will act based on the previous robot action, and vice versa.

We define $\boldsymbol{\theta}$ as the set of $k$ representative transition matrices $\theta_{1},...,\theta_{k}$ that correspond to the $k$ clusters. Every sequence $x_{i}$, each of length $l$, in the data $D={x_1...x_n}$ must be assigned to one of these $k$ clusters. The assignments of these sequences to clusters can be denoted as $Z={z_1...z_n}$, where each $z_{i} \in \{1,...,k\}$ . 

The probability of one sequence $x_{i}$ parameterized by $\boldsymbol{\theta}$ can be represented as follows:
\begin{equation}
\label{e:estimate}
\begin{aligned}
P(x_{i}; \boldsymbol{\theta}) &= \sum\limits_{z_{i}=1}^{k}P(z_{i})P(x_{i}|z_{i}; \boldsymbol{\theta}) 
 \\ &= \sum\limits_{z_{i}=1}^{k}P(z_{i})\left(\prod\limits_{j=2}^{l} \theta_{z_{i}}(x_{i}^{j}| x_{i}^{j-1})\right)
\end{aligned}
\end{equation}
$x_{i}^{j}$ denotes the $j$\textsuperscript{th} element of the $i$\textsuperscript{th} demonstrated sequence. 

For all data points, the log-likelihood can be represented as follows:
\begin{equation}
\label{e:loglikelihood}
\begin{aligned}
l(D; \boldsymbol{\theta}) &= \sum\limits_{i=1}^{n}log P(x_i; \boldsymbol{\theta})
\\ &= \sum\limits_{z=1}^{k}\sum\limits_{i=1}^{n}\delta(z|z_{i})log\left(P(z_{i}) \prod\limits_{j=2}^{l} \theta_{z_{i}}(x_i^{j}| x_i^{j-1})\right)
\end{aligned}
\end{equation}
\quad $\delta(z|z_{i}) = 1$ if $z = z_{i}$ and zero otherwise.


\begin{figure}
\centering
\label{fig:EM_algo}
\begin{center}
\linethickness{.4mm}
{\line(1,0){240}}
\end{center}
\vspace{-1.5mm}
{\bf Algorithm:} Cluster-Transition-Matrices ($k$)
\begin{enumerate}
  \item Initialize $\boldsymbol{{\hat{\theta}}}$ by randomizing $\hat{\theta}_{1},...,\hat{\theta}_{k}$
  \item Initialize sequence assignments $Z=z_{1},...,z_{n}$
  \item {\bf repeat} 
  \item \quad {\em E-step}: Compute assignments for each sequence $z_{i}$
  
  \quad for  $i = 1, ... , n$
  

  \quad $z_{i} = \arg\max\limits_{z_{i}} \left(\prod\limits_{j=2}^{l} \hat{\theta}_{z_i}(x_i^{j}| x_i^{j-1})\right)$
  
  \item \quad {\em M-step}: Update each transition matrix $\hat{\theta}_{z}$ 
  
  \quad for $z=1,...,k$
  
  \quad $n_{i|j}$ : observed count of transitions from $i$ to $j$
  
  \quad $\hat{\theta}_{z, i|j} = \frac{n_{i|j}} {\sum\limits_{x=1}^{|A|} n_{x|j}}$ for $i,j=1,...,|A|$ 

\item {\bf until} $Z$ converges to stable assignments
\end{enumerate}
\vspace{-1.5mm}
\begin{center}
\linethickness{.4mm}
{\line(1,0){240}}
\end{center}
\vspace{-1.5mm}
\caption{Cluster Transition Matrices using EM}
\end{figure}

\begin{figure}
\centering
\label{fig:BIC_algo}
\begin{center}
\linethickness{.4mm}
{\line(1,0){240}}
\end{center}
\vspace{-1.5mm}
{\bf Algorithm:} Select-Best-Model ($k_{min}$, $k_{max}$, $numOfIterations$)
\begin{enumerate}
\item {\bf for} $k=k_{min}$ {\bf to} $k_{max}$

\item \quad {\bf for} $i=0$ {\bf to} $numOfIterations$

\item \quad \quad Call Clustering-Transition-Matrices-using-EM ($k$)

\item \quad \quad Calculate the log-likelihood for this model:

\quad \quad $l(D; \boldsymbol{\hat{\theta}}) = $

\quad \quad \quad $\sum\limits_{z=1}^{k}\sum\limits_{i=1}^{n}\delta(z|z_{i})log\left(P(z_{i}) \prod\limits_{j=2}^{l} \theta_{z_{i}}(x_i^{j}| x_i^{j-1})\right)$ 

\item \quad \quad Calculate BIC term for this value of $k$: 

\quad \quad $BIC = l(D; \boldsymbol{\hat{\theta}}) - \frac {K}{2} log(n)$\\

\quad \quad where $K$ is the number of parameters 

\quad \quad and $n$ is the number of data points.

\item \quad For the current value of $k$, choose the cluster 

\quad partition with the highest BIC value.

\item Return the value of $k$ with the maximum BIC value and the corresponding cluster partition.

\end{enumerate}
\vspace{-1.5mm}
\begin{center}
\linethickness{.4mm}
{\line(1,0){240}}
\end{center}
\vspace{-1.5mm}
\caption{Finding Ideal Number of Clusters using BIC}
\end{figure}

The Cluster-Transition-Matrices EM algorithm learns the optimal transition matrices $\hat{\theta}_{1},...,\hat{\theta}_{k}$ by iteratively performing the E-step and the M-step. First, lines 1-2 randomly initialize $k$ transition matrices and sequence assignments; then, lines 3 through 6 repeatedly execute the E-step and M-step until the assignments $Z$ converge to stable values. In the E-step, we complete the data by assigning each sequence to the cluster with the highest log-likelihood (line 4). In the M-step, each cluster's transition matrix is updated by counting the transitions in all sequences assigned to that cluster (line 5). These two steps are repeated until the assignments $z_{1},...,z_{n}$ do not change (line 6).

The EM algorithm used here requires $k$ as input, but since we use an unsupervised clustering approach, this value is unknown to us. We run the Select-Best-Model algorithm to find the ideal value of $k$ using Bayesian information criterion (BIC). We specify the range of possible values for $k$: $k_{min} - k_{max}$ as input, and run EM for each value of $k$ within this range. For our problem, we chose the range for $k$ to be from 2 to 10, due to the observation that there tends to be only a few high-level human preferences for a particular task (line 1). In addition to testing multiple values of $k$, we run multiple iterations of EM for each value of $k$, as specified by the input $numOfIterations$, because the results can differ according to initialization and EM often finds locally optimal solutions. In our case, we use $numOfIterations = 20$, as this is sufficient to see consistent results (line 2). After each run of EM, we calculate the log-likelihood based on the resulting cluster partition, as specified in line 4. We then use BIC to introduce a penalty term for complex models, so that the penalty increases with the value of $k$. In this case, the number of parameters $K$ is $k|A|(|A|-1)$, since we have $k$ transition matrices, each of which has $|A|(|A|-1)$ free parameters (line 5). The cluster partition with the highest BIC value over all the iterations was chosen as the best model for that particular value of $k$ (line 6). Comparing the BIC values for each value of $k$ then determines the final cluster partition (line 7). By using EM and BIC in this way, we can find both the number of clusters and the cluster partition for this data.

We then input the learned clusters into a Mixed-Observability Markov Decision Process, which treats the human type as a partially observable variable that can take a finite set of values. Each human type value is associated with a corresponding cluster. In the next section, we describe the MOMDP formulation, the learning of a reward function for each human type value and the computation of an approximately optimal policy for the robot.


\section{Mixed Observability Markov Decision Process Learning and Planning}

The clusters of the demonstrated action sequences represent different types of humans. When a robot is introduced to work with a new human worker, it needs to infer the human type for that worker and choose actions that are aligned to their preference. Additionally, the robot should reason over the uncertainty on the type of the human. Therefore, the cluster indices serve as the values of a partially observable variable denoting the human type in a Mixed-Observability Markov Decision Process. Our framework learns a reward function for each human type, which represents the preferences of a human of the given type for a subset of task-related robot actions. We then compute an approximately optimal policy for the robot that reasons over the uncertainty on the human type and maximizes the expected accumulated reward. 

We describe the MOMDP formulation, the learning of the reward function and the computation of an approximately optimal policy as follows:

\subsection{MOMDP Formulation} \label{sec:MOMDPformulation}
%
%

We treat the unknown human type as a hidden variable in a Mixed-Observability Markov Decision Process (MOMDP), and have the robot choose actions according to the estimated human type. The MOMDP framework uses proper factorization of the observable and unobservable state variables, reducing the computational load. The MOMDP is described by a tuple, $\{X, Y, S, A_{r}, \mathcal{T}_{x}, \mathcal{T}_{y}, R,\Omega, O\}$, so that:\begin{itemize}

\item ${X}$ is the set of observable variables in the MOMDP. In our framework, the observable variable is the current task-step among a finite set of task-steps that signify progress toward task completion. 

\item ${Y}$ is the set of partially observable variables in the MOMDP. In our framework, a partially observable variable, $y$, represents the human type. 


\item $S:X \times Y$ is the set of states in the MOMDP consisting of the observable and non-observable variables. The state $s \in S$ consists of the task-step $x$, which we assume is fully observable, and the unobservable type of the human $y$. 

\item $A_{r}$ is a finite set of discrete task-level robot actions.


\item $\mathcal{T}_{x}: S  \times A_{r} \longrightarrow \Pi(X)$ is the probability of the fully observable variable being $x'$ at the next time step if the robot takes action $a_{r}$ at state $s$. 

\item $\mathcal{T}_{y}: S  \times A_{r} \times X \longrightarrow \Pi(Y)$ is the probability of the partially observable variable being $y'$ at the next time step if the robot takes action $a_{r}$ at state $s$, and the next fully observable state variable has value $x'$. 

\item $R :  S \times A_{r} \longrightarrow \mathbb{R}$ is a reward function that gives an immediate reward for the robot taking action $a_{r}$ at state $s$. It is a function of the observable task-step $x$, the partially observable human type $y$ and the robot action $a_{r}$. 

\item $\Omega $ is the set of observations that the robot receives through observation of the actions taken by the human and the robot. 
\item $O : S \times A_{r} \longrightarrow \Pi(\Omega)$ is the observation function, which gives a probability distribution over possible observations for each state $s$ and robot action $a_{r}$. We write $O(s, a_{r}, o )$ for the probability that we receive observation $o$ given $s$ and $a_{r}$.
\end{itemize}

\subsection{Belief-State Estimation}

Based on the above, the belief update is then \cite{ong2010planning}:
\begin{equation}
\label{e:beliefupdate}
\begin{aligned}
b_{y}(y') = \eta O(s', a_{r}, o) \sum_{y \in Y}{\mathcal{T}_{x}(s, a_{r}, x') \mathcal{T}_{y}(s,a_{r},s')b_{y}(y)}
\end{aligned}
\end{equation}

\subsection{Inverse Reinforcement Learning}

Given a reward function, an exact value function and an optimal policy for the robot can be calculated. Since we want the robot to choose actions that align with the human type of its teammate, a reward function must be specified for every value that the human type can take. Manually specifying a reward function for practical applications can be tedious and time-consuming, and would represent a significant barrier in the applicability of the proposed framework. In this section, we describe the learning of a reward function for each human type using the demonstrated sequences that belong to the cluster associated with that specific type. 

For a fixed human type $y$, we can reduce the MOMDP into a Markov Decision Process (MDP). The Markov Decision Process in this context is a tuple: $(X, A_{r}, T_{x}, R, \gamma)$, where $X$, $A_{r}$, $T_{x}$ and $R$ are defined in the \nameref{sec:MOMDPformulation} section above. Given demonstrated sequences of state-action pairs, we can estimate the reward function of the Markov Decision Process using the Inverse Reinforcement Learning (IRL) algorithm \cite{Abbeel04}. Note that we assume the human type to be constant in the demonstrated sequences. 
To compute the reward function for each cluster, we first assume that a feature vector $\varphi$ exists for each state, and each given policy has a feature expectation that represents the expected discounted accumulation of feature values based on that policy. Formally, we define the feature expectations of a policy $\pi$ to be:  

\begin{equation}
\label{e:featureExp}
 \mu (\pi)=E[\sum_{t=0}^{\infty } \gamma^{t}\varphi (s_{t})|\pi]
\end{equation}

We require an estimate of the feature expectations for each human type. Given a set of $n_z$ demonstrated state-action trajectories per human type $z$, we denote the empirical estimate for the feature expectation as follows:

\begin{equation}
\label{e:expertFeatureExp}
\hat{\mu}_{z} =\frac{1}{n_z}\sum_{i=1}^{n_z}\sum_{t=0}^ {\infty}\gamma^{t}\varphi(s_{t}^{(i)})
\end{equation}

The IRL algorithm begins with a single random policy and attempts to generate a policy that is a mixture of existing policies, with feature expectations that are similar to those for the policy followed by the expert. In our case, the ``expert" demonstrations were the demonstrations followed by all humans of a particular type. The algorithm terminates when $||\mu_z - \mu(\tilde\pi)||_2 \leq \epsilon$, and is implemented as follows:

\begin{enumerate}
  \item Randomly pick some policy $\pi^{(0)}$ and approximate via Monte-Carlo $\mu^{(0)} = \mu(\pi^{(0)})$, and set $i = 1$.
  \item Compute a new ``guess" of the reward function by solving the following convex (quadratic) programming problem:
	\begin{equation}
	\label{e:quadraticProg}
	\min_{\lambda, \mu} ||\hat\mu_z - \mu||_2
	\end{equation}
  subject to $\sum_{j=0}^{i-1}\lambda_j \mu^{(j)} = \mu$, $\lambda \geq 0$, and $\sum_{j=0}^{i-1} \lambda_j = 1$. We set $t^{(i)} = ||\hat\mu_z - \mu||_2$ and $w^{(i)} = \frac{\hat\mu_z - \mu}{ ||\hat\mu_z - \mu||_2}$.
  \item If $t^{(i)} \leq \epsilon$, then terminate. 
  \item Use reinforcement learning to compute the optimal policy $\pi^{(i)}$ of the MDP with the reward function $R(s)=w^{(i)}\phi(s)$.
  \item Approximate via Monte-Carlo $\mu^{(i)} = \mu(\pi^{(i)})$.
  \item Set $i = i+1$, and go back to step 2.
\end{enumerate}
The result of the IRL algorithm is a list of policies, $\{\pi^{(i)}:i=0\ldots M\}$, with weights $\lambda_i$ and feature counts $\mu^{(i)}$, where $M$ is the total number of iterations. Each policy $i$ maximizes the expected accumulated reward function, calculated as $R(s)=w^{(i)}\phi(s)$. The output of the algorithm is a reward function, computed by taking the mean of the weight values $w^{(i)}$ over the second half of the iterations. We ignore the first half for the calculation, as the initial policies and associated weights generated by the algorithm are of lesser quality.

For each human type, the framework applies inverse reinforcement learning, using the demonstrated sequences of that type as input to calculate an associated reward function. With a reward function for any assignment of the partially observable human type variable $y$, we can now compute an approximately optimal policy for the robot, as described in the next section. 
\subsection{Policy Computation}

We solve the MOMDP for a policy that takes into account the uncertainty of the robot over the human type, while maximizing the agent's expected total reward. MOMDPs are structured variants of POMDPs, and finding an exact solution of a POMDP is computationally expensive \cite{Kaelbling199899}. Point-based approximation algorithms have greatly improved the speed of POMDP planning ~\cite{shani2013survey, kurniawati2008sarsop, pineau2003point} by updating selected sets of belief points. In this work we use the  SARSOP solver~\cite{kurniawati2008sarsop} which, combined with the MOMDP formulation, can scale-up to hundreds of thousands of states~\cite{bandyopadhyay2013intention}. The SARSOP algorithm samples a representative set of points from the belief space that are reachable from the initial belief, and uses this set as an approximate representation of the space, allowing for the efficient computation of a satisfactory solution.

\section{Evaluation}
In this section, we show the applicability of the proposed framework to two separate applications of human-robot teamwork: a place-and-drill task and a hand-finishing task.

For the place-and-drill task, we used data collected from a human subject experiment in which 18 human subjects provided demonstrations for a shared-location joint-action collaborative task. The role of the human was to place screws in one of three available positions, while the robot was to drill each placed screw. The demonstrations were provided during a training phase in which the human and robot switched roles, giving the human the opportunity to demonstrate robot drilling actions to show the robot how he would like the task to be executed. 

To evaluate our framework, we used leave-one-out cross-validation, by removing one subject and using the demonstrated sequences from the remaining 17 subjects as the training set. We applied the clustering algorithm presented in the section \nameref{sec:clustering}. In all cross-validation iterations, the human subjects were clustered into two types: a ``safe" type, in which each screw was placed before drilling began, and an ``efficient" type, in which each screw was drilled immediately after placement. The order of screw placement did not affect the clustering. 

For each type, our framework uses the inverse reinforcement learning algorithm to learn a reward function associated with that type. The number of types and their associated reward functions is then passed to the MOMDP formulation as input. Therefore, the unobservable variable $y$ in the MOMDP formulation could take two values: ``safe" or ``efficient". The observable state variable for this task, $x$, is the workbench configuration. The actions $a_{r}$ are the drilling actions, as well as the no-op action. The human placing actions are encoded implicitly in the transition and observation matrices. 

\begin{figure}[h]
\centering
 {\includegraphics[width=1.0\linewidth]{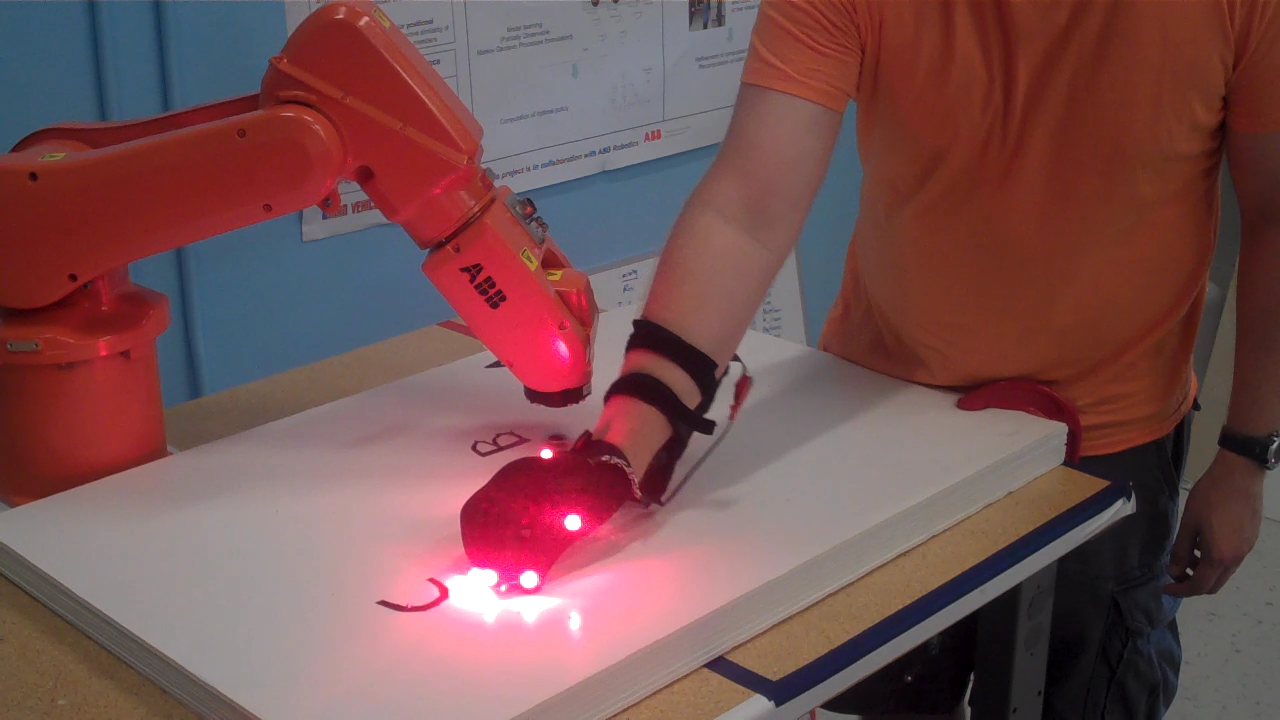}}\hfil
 \caption{Task execution by a human-robot team on a place-and-drill task. }

 \label{fig:TaskExecution}
\end{figure}

Each subject left out of the training set for cross-validation - referred to as the ``testing subject" - provided three demonstrated sequences of human and robot actions and a probability distribution over its type was calculated according to Equation~\ref{e:estimate}. Using this as the initial belief on the human type, and the associated reward function from the inverse reinforcement learning algorithm, the SARSOP solver computed a policy for the robot.  We then had the testing subject execute the place-and-drill task with the actual robot, with each performing their predefined roles, during the ``task execution phase" (Figure~\ref{fig:TaskExecution}). We assumed that the preference of the testing subject for the robot actions was the same when both demonstrating the actions and executing the task; this assumption was later verified by subject responses to a post-experimental questionnaire.


\subsection{Performance of Human Type Clustering}

Clustering of human types was performed on the place-and-drill data, which consisted of 54 demonstrated sequences (3 for each of the 18 participants). We considered only data collected during the phase in which the human performed the robot actions and the robot performed the human actions, as this phase revealed the human's preference for the robot actions. 

To evaluate whether the robot was acting as a human would, we compared the order of actions when the robot performed as a human with the order in which the human performed them, and found that the actions matched 88\% of the time. This indicates that, although the robot did not perform exactly as the human did, the orders of actions were similar enough that we can use these sequences to test our clustering approach. 

To validate the clustering for this data, we used leave-one-out cross-validation, with testing on each participant. Because each participant had 3 demonstrated sequences, we took a weighted average of the resulting assignments of these 3 sequences to determine the final type assigned to each participant. The weights were obtained from the likelihoods calculated via the Cluster-Transition-Matrices EM algorithm. The type predicted by the clustering algorithm was then compared with manual labels hand-coded by a human expert without knowledge of the clustering algorithm. Two human types were identified by the clustering algorithm and the human expert, labeled ``safe"  and ``efficient" by the human expert. The ``safe" type of person prefered that the robot wait until all screws were placed by the person before drilling. The  ``efficient" type of person prefered to place and drill in alternating fashion to finish the task quickly. We obtained an average classification accuracy of 96.5\% for this data. This algorithm performed well on this data set due to a clear partition in separation distance between the two types. 

\subsection{Robustness of Computed Policy}
We compared the computed policy with a state-of-the-art iterative algorithm for human-robot collaborative tasks, called "human-robot cross-training" \cite{Nikolaidis2013}, in which the robot learns a human model by switching roles with the human. We used the demonstrated sequences of the testing subject as input for the cross-training algorithm, which learns a user model by updating the transition and reward function of a Markov Decision Process. The algorithm then computes a policy based on the learned model, which matches the human preference during task execution when the human and robot resume their predefined roles \cite{Nikolaidis2013}.

In the actual human subject data, the human placement actions during task execution were, in most cases, identical to those provided during the demonstrations. Therefore, we simulated the task execution for increasing degrees of deviations from the demonstrated actions of the human. For instance: If, in the demonstrated sequences, the human placement actions were to first, ``place screw A"; second, ``place screw B", and finally, ``place screw C", during actual task execution we gradually increased the probability of the human choosing a different placement action at each task-step, leading the execution to previously unexplored parts of the state-space. We did this by having a simulated human perform a random placement action with a probability $\epsilon$, or the actual action taken by the testing human subject with probability $1-\epsilon$. The x-axis of Figure~\ref{fig:MeanRewards} denotes the value of $\epsilon$.

For increasing levels of deviations, we computed the accumulated reward for the policy of the proposed framework and the policy computed by the human-robot cross-training algorithm. We did this for each iteration of the cross-validation, and plotted the mean accumulated reward (Figure~\ref{fig:MeanRewards}).  The policy of the human-robot cross-training algorithm performed similarly to the one of the proposed framework if the user did not deviate from his demonstrated placing actions. However, as the deviations increased, the policy from the cross-training algorithm performed worse. On the other hand, the MOMDP agent reasons over the partially observable human type using a reward function that learns from all demonstrated sequences that belong to the cluster associated with that type; therefore, its performance was not affected by these deviations.

\subsection{Quality of Learned Model}
\begin{figure}[h]
\centering
 {\includegraphics[width=1.0\linewidth]{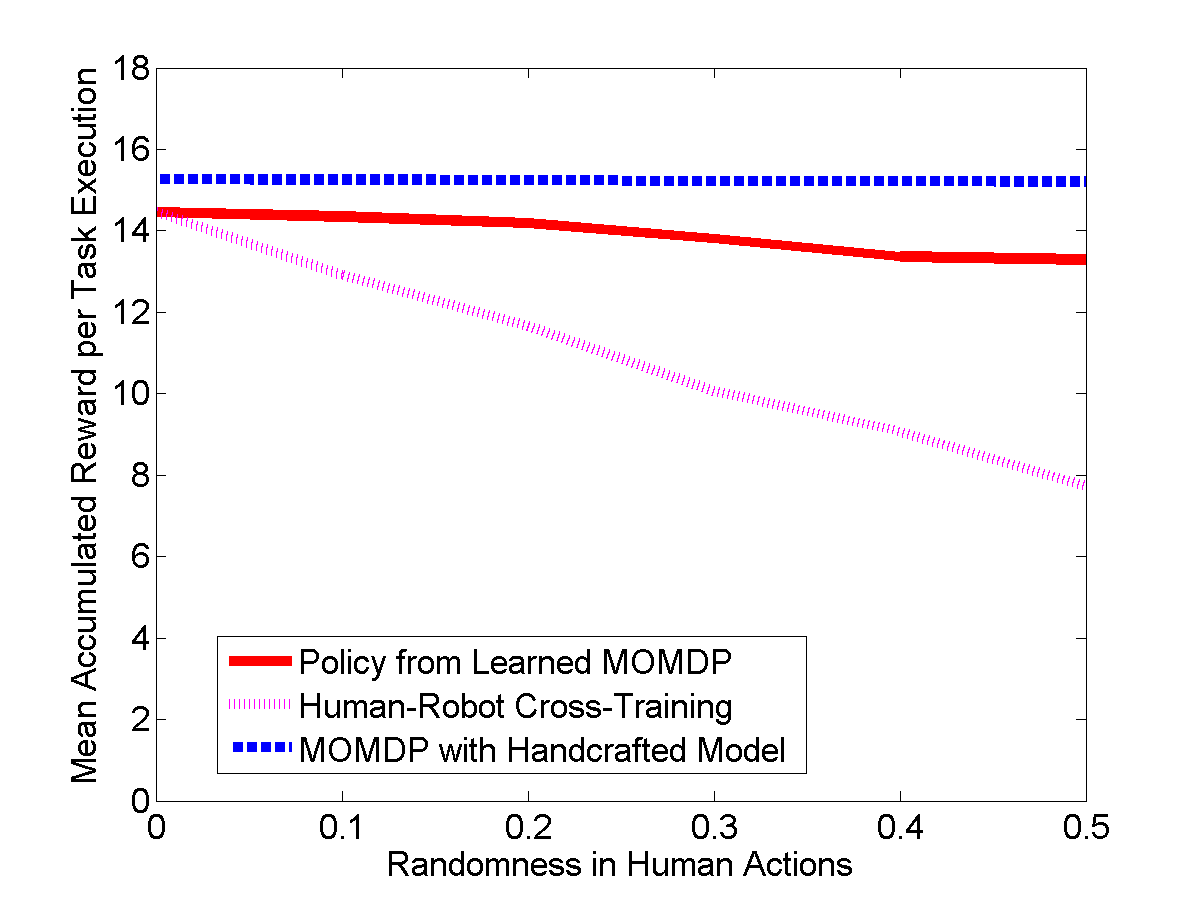}}\hfil
 \caption{Accumulated reward averaged over 18 iterations of cross-validation (one for each human subject). The plotted lines illustrate the performance of a policy of a MOMDP model hand-coded by a domain expert, the learned policy of the automatically generated MOMDP model using the proposed framework and the learned policy from the Human-Robot Cross-Training algorithm. The x-axis represents the probability of the human performing a random action instead of replaying the action he actually took during the task-execution phase with the robot. For each subject, we ran 100 simulated iterations of task execution. }
\label{fig:MeanRewards}
\end{figure}

To evaluate the quality of the clusters and corresponding reward functions generated automatically from our framework, we had a domain expert manually partition the data and empirically hand-craft a reward function. Figure~\ref{fig:MeanRewards} shows that the policy computed by the MOMDP using the automatically generated user model has comparable performance to the one that uses the hand-coded model. The plotted lines denote the accumulated reward, averaged over all iterations of cross-validation. 

\subsection{Online Estimation of Human Type} 

If the human actions are informative of his type \textemdash his preference for the actions of the robot \textemdash the human type can be estimated online. We demonstrate this capability by using an MOMDP formulation on a large-scale hand-finishing task. In this task, the role of the robot is to position and orient a large box, while the role of the human is to refinish the two side surfaces of the box. The box is attached to the robot end-effector, and the robot can move the box along the horizontal and vertical axis and rotate it in the tilt direction. The human can choose to first refinish the left and then the right surface of the box, or vice versa. The goal is for robot to estimate the preference of the human on the ordering of box rotations, so that it can move the box to a reachable, ergonomically friendly position.

The observable state variables $x$ of the MOMDP framework are the box position, which takes discrete values in the horizontal and vertical axis, and the box tilt angle. Demonstrated data from 6 subjects was run by the Cluster-Transition-Matrices EM algorithm. The output of this algorithm was two clusters that indicated the user's preference of the order of refinishing (the left side first vs. the right side). These are the values of the unobserved state variables $y$. The total size of the state-space was 2000 states. The human actions were estimated by tracking the 3D position of the human hand via the Phasespace motion-capture system~\cite{phasespace}. In the MOMDP formulation, we define the observation function $O$ as a discretized 2D-Gaussian, giving the probability of the position of a human hand in the horizontal plane conditioned on the human preference on the ordering (left side first vs. right side first). The mean and variance were empirically specified in this work, and we leave learning the observation function from human team demonstrations for future work. 

In this application we perform online estimation of the human type based on observations, rather than using offline demonstrations from the user. Figure~\ref{fig:HandFinishingGraphs} illustrates an execution of the task by a human robot team. The person starts moving towards the right side of the robot, and therefore the estimate that human prefers the robot to present the right side of the box first increases. However, the robot waits to gain more information before starting to move the box. The person then walks towards the robot and slightly to the left. The robot updates its belief, and when the uncertainty over the human type is low enough, the robot moves the box to the left and rotates it, so that the person can do the refinishing of the left-surface. This emulates behavior frequently seen in human teams, wherein a human worker introduced with a new teammate may stand and wait to see what his co-worker will do, associate his partner's actions with previously learned behaviors, and move on with the task execution after having enough confidence on his teammate's intentions.

\begin{figure}[h]
\centering
 \subfigure[]
    {
        \includegraphics[width=1.0\linewidth]{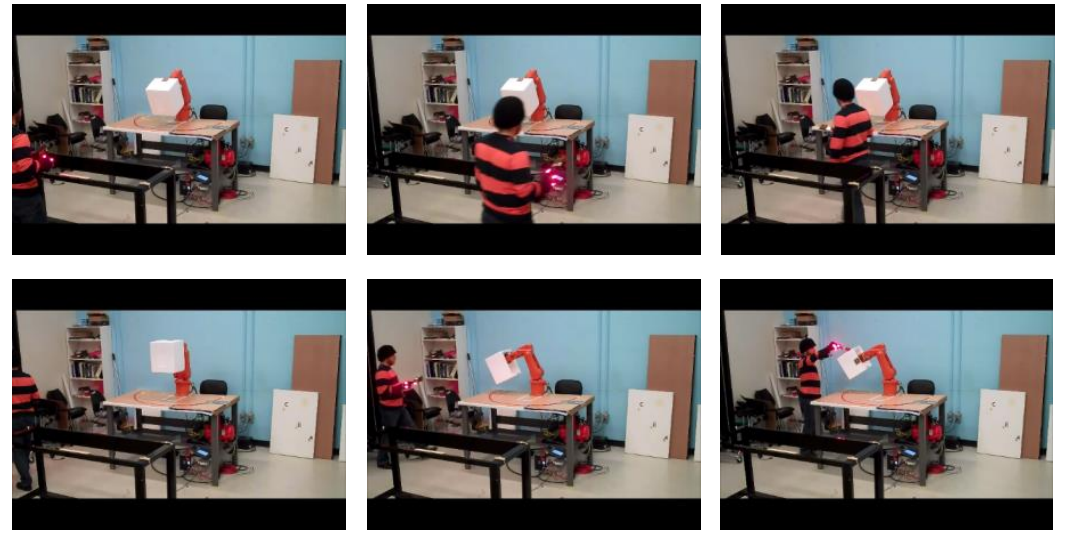}
        \label{fig:snapshots}
    }
    \\
    \subfigure[]
    {
        \includegraphics[width=1.0\linewidth]{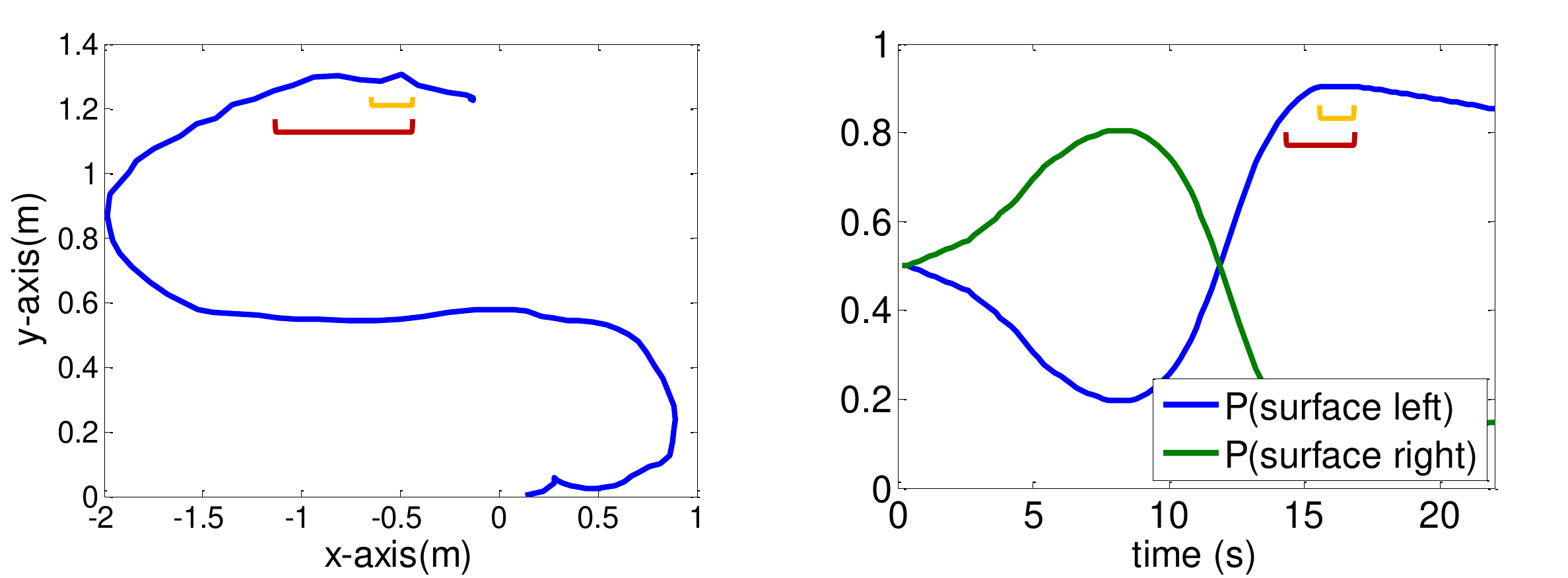}
        \label{fig:HandFinishingGraphs}
    }
 \caption{(a) Execution of a hand-finishing task by a human worker and an industrial robot. The worker initially moves to the right (top-center), but the robot remains still. When the human worker moves to the left side of the room (bottom-left) the robot moves the box to the left, so that is reachable by the human (bottom-center). Then, the robot rotates the box, so that the human can refinish the surface( bottom-right). (b) The plot on the left shows the human trajectory in the horizontal plane (top view). The plot on the right shows the belief for the human preference, as it evolves with time. The red bracket in both plots illustrates the time-period when the robot is moving the box to the left, and the orange bracket the time period when the robot is rotating the box in the tilt direction. }
 \label{fig:HandFinishingGraphs}
\end{figure}


\section{Conclusion}
We have presented a framework that automatically learns the ``dominant" types of human subjects when working in teams on a collaborative task. Assuming access to a set of demonstrated sequences of  actions from human teams, we find the number of human types by clustering these sequences. We then learn a user model for each type, represented by a reward function of a Mixed Observability Markov Decision Process. An approximately optimal policy that maximizes the expected accumulated reward is computed, taking into consideration the uncertainty on the human types. When a new human subject is introduced to execute the collaborative task with the robot, his type is inferred either offline from prior demonstrations or online during task execution. 

Evaluation showed that the robot performance was robust to increasing deviations of human behavior from the demonstrated actions, compared to previous algorithms that reason in state-space. Furthermore, the performance was comparable to the policy of a MOMDP agent computed using a hand-coded model by a domain expert. These results indicate that models of human types in collaborative tasks can be efficiently learned and integrated into general decision-making, enabling robots to develop robust policies aligned with the personalized style of their human partners. 

\bibliographystyle{plainnat}
\bibliography{mybib2}

\end{document}